\documentclass[journal]{IEEEtran}

\usepackage[pdftex]{graphicx}
\usepackage{amsmath}
\usepackage{bm}
\usepackage{stfloats}
\usepackage{subfigure}
\usepackage{cite}
\usepackage{multirow}

\begin{document}

\title{\LARGE A Deep Learning Based Attack for The Chaos-based Image Encryption}

\author{Chen~He,~\IEEEmembership{Member,~IEEE,}
        Kan~Ming,
        Yongwei~Wang,~\IEEEmembership{Student Member,~IEEE,}   
        and~Z.~Jane~Wang,~\IEEEmembership{Fellow,~IEEE}}

\maketitle


\begin{abstract}
In this letter, as a proof of concept, we propose a deep learning-based approach to attack the chaos-based image encryption algorithm in \cite{guan2005chaos}. The proposed method first projects the chaos-based encrypted images into the low-dimensional feature space, where essential information of plain images has been largely preserved. With the low-dimensional features, a deconvolutional generator is utilized to regenerate perceptually similar decrypted images to approximate the plain images in the high-dimensional space. Compared with conventional image encryption attack algorithms, the proposed method does not require to manually analyze and infer keys in a time-consuming way. Instead, we directly attack the chaos-based encryption algorithms in a key-independent manner. Moreover, the proposed method can be trained end-to-end. Given the chaos-based encrypted images, a well-trained decryption model is able to automatically reconstruct plain images with high fidelity. In the experiments, we successfully attack the chaos-based algorithm \cite{guan2005chaos} and the decrypted images are visually similar to their ground truth plain images. Experimental results on both static-key and dynamic-key scenarios verify the efficacy of the proposed method.  

\end{abstract}

\begin{keywords}
 chaos-based encryption, image decryption, deep learning.
\end{keywords}


\section{Introduction}

A series of characteristics of chaotic systems such as pseudo-random characteristics, unpredictability of orbit, sensitivity to initial state and control parameters are in good agreement with many requirements of cryptography, so chaotic cryptography has been extensively studied. Whit this, a chaos-based image encryption algorithm was proposed in \cite{guan2005chaos}. The algorithm combines confusion and diffusion in traditional cryptography, which utilizes Arnold's cat map\cite{peterson1997arnold} to shuffle the positions of plain-image pixels to introduce diffusion and Chen's chaotic system \cite{chen1999yet} to change the grayscale values of the shuffled image pixels to introduce confusion, the combined transformation of confusion and diffusion provides greater security than using them separately.

In order to ensure the security level of the encryption algorithm, the researchers also continue to analyze the vulnerabilities of various encryption schemes and try to attack them. Among them, a traditional method is to infer the key by manually analyzing the encryption algorithm, just like the solution in \cite{cokal2009cryptanalysis}, the authors demonstrated a chosen-plaintext attack and a known-plaintext attack that reveals the secret parameters of the encryption algorithms in \cite{guan2005chaos}, however this method is key-dependent and relatively time-consuming and labor intensive. Another approach is to search for keys from a key dictionary maintained with some special algorithms. For instance, in \cite{hitaj2017passgan} a key dictionary was constructed by using machine learning algorithms to generate more human-compliant keys, the attacker constantly searches through the key dictionary until the correct one is found, but in addition to being time consuming, this solution has a great chance of not finding the right key.

In this letter, a novel image decryption approach is proposed to attack the chaos-based image encryption algorithm \cite{guan2005chaos} based on deep learning. As proof of concept, we first proposed to use the deep learning method to crack the encryption algorithm. First, we extract essential features from cipher-images with a convolutional encoder architecture. Then we regenerate decrypted images from the features to approximate their corresponding plain images with a symmetric deconvolutional generator network. Experimental results demonstrate the effectiveness of the proposed method both for cracking the proposal \cite{guan2005chaos} in both static-key and dynamic-key encryption cases. Compared to previous encrypted image attack schemes \cite{cokal2009cryptanalysis, hitaj2017passgan}, our method does not require a time-consuming manual analysis of the algorithm to infer the key. Instead, we directly attack the chaotic-based encryption algorithm in a key-independent manner. After training the decryption model, the well-trained model can quickly and automatically reconstruct the plain image from the encrypted image with high fidelity.

\begin{figure*}[ht!] 
\centering
\includegraphics[width=1\textwidth]{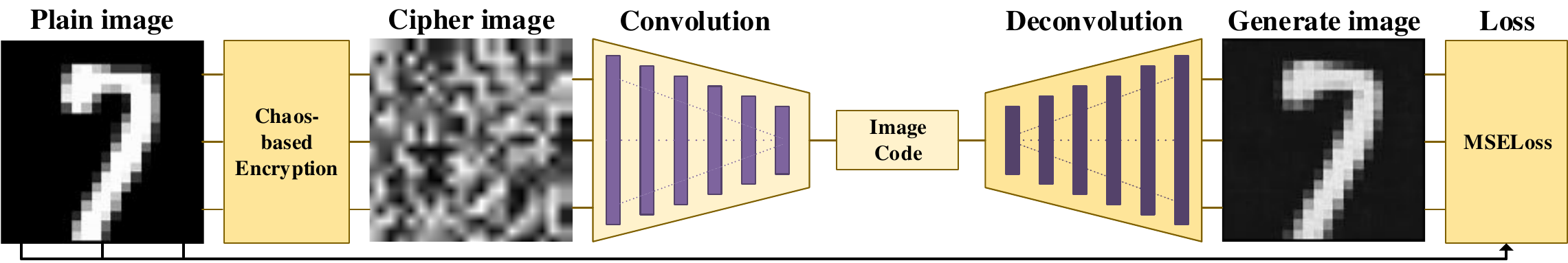} 
\caption{Illustration of the proposed deep learning-based image decryption scheme for the chaos-based image encryption method \cite{guan2005chaos}. The proposed method enables end-to-end relationship inference between plain images and chao-based encrypted images. The decryption model can be divided into two parts: the first part is to project cipher images into the low-dimensional feature space with a convolutional encoder, and the second part is to regenerate the decrypted image with a deconvolutional generator.}
\label{model}
\end{figure*}


\section{The chaos-based image encryption algorithm}
There are two steps in the chaos-based image encryption algorithm \cite{guan2005chaos}    : first, the positions of image pixels in the plain images are shuffled using Arnold's cat map \cite{peterson1997arnold}. Then, the grayscale values of the shuffled image pixels are changed using Chen's chaotic system \cite{chen1999yet}.


\subsection{Arnold's cat map}
Arnold's cat map \cite{peterson1997arnold} is a chaotic mapping method for repeated folding and stretching transformation in a finite region, which is generally applied to multimedia chaotic encryption. Without loss of generality, assume that we have an original grayscale image $P$ of size $N\times N$ with pixel coordinates $S=\left\{(x,y)|x,y=0,1,2,...,N-1\right\}$. Arnold's cat map can be expressed as, 
\begin{equation}
    \begin{bmatrix}
    x' \\
    y'
    \end{bmatrix}
    =\mathbf{A}
    \begin{bmatrix}
    x \\
    y
    \end{bmatrix}\left(\bmod N \right)
    =\begin{bmatrix}
    1 & p \\
    q & pq + 1
    \end{bmatrix}
     \begin{bmatrix}
    x \\
    y
    \end{bmatrix}\left(\bmod N\right),
\end{equation}
where $p$ and $q$ are positive integers. $\left(x',y'\right)$ is the new coordinate value of the original pixel $\left(x,y\right)$ after iterating the map once. After Arnold's cat map has been performed for $n$ times, we have,
\begin{equation}
    \begin{bmatrix}
    x' \\
    y'
    \end{bmatrix}
    =\mathbf{A}^n
    \begin{bmatrix}
    x \\
    y
    \end{bmatrix}\left(\bmod N\right)
    =\mathbf{M}
     \begin{bmatrix}
    x \\
    y
    \end{bmatrix}\left(\bmod N\right),
\end{equation}
where
\begin{equation}
    \mathbf{M} = 
    \begin{bmatrix}
    m_1 & m_2 \\
    m_3 & m_4 
    \end{bmatrix}
    = \mathbf{A}^n \left(\bmod N\right).
\end{equation}


\subsection{Chen's chaotic system}

In Chen's chaotic system \cite{chen1999yet}, there are a set of differential equations given as,
\begin{equation}
      \left\{
             \begin{array}{lr}
             \dot{x}=a(y-x)       \\
             \dot{y}=(c-a)x-xz+cy \\
             \dot{z}=xy-bz,  
             \end{array}
      \right.
\end{equation}
where $a$, $b$, and $c$ are parameters. The system is chaotic when $a=35$, $b=3$ and $c\in\left[20,28.4\right]$ \cite{chen1999yet,guan2005chaos}. 


\subsection{The Chaos-based encryption algorithm}
In \cite{guan2005chaos}, the secret keys of the encryption algorithm are $p$, $q$, the number of iterations $n$ of Arnold's cat map and the initial value of Chen's chaotic system, i.e., $x_0$, $y_0$, $z_0$. The specific steps are as follows:
\\(1) Obtain the shuffled image $S$ by using Arnold's cat map to shuffle the image $P$.
\\(2) Get a pixels sequence $S = \left\{s_1, s_2, ..., s_{N\times N}\right\}$ by scanning the shuffled image $S$ in order from left to right and then top to bottom.
\\(3) Iterate Chen's chaotic system $N_0 = (N\times N) / 3$ times by using Runge-Kutta step size 0.001, in each iteration, we can get three values  $x_i$, $y_i$ and $z_i$,$1 \leq i \leq N_0$, by processing these values as follows: 
\begin{equation}
    \begin{array}{lr}
       k_{3(i-1)+1} = (\left|x_i\right|
                      -\lfloor\left|x_i\right|\rfloor)
                      \times 10^{14} \bmod 256,\\
       k_{3(i-1)+2} = (\left|y_i\right|
                      -\lfloor\left|y_i\right|\rfloor)
                      \times 10^{14} \bmod 256,\\
       k_{3(i-1)+3} = (\left|z_i\right|
                      -\lfloor\left|z_i\right|\rfloor)
                      \times 10^{14} \bmod 256,
    \end{array}
\end{equation}
the encryption key sequence $K = \left\{k_1, k_2, ..., K_{N\times N}\right\}$ will be obtained, where $\left|x\right|$ is the absolute value of $x$. $\lfloor x \rfloor$ means round down, it returns the largest integer not larger than $x$. In the cryptosystem, all variables have a 15-digit precision when expressed in scientific notation, so the decimal fractions of the variables need to be multiplied by $10^{14}$.
\\(4) Encrypt the shuffled sequence $S = \left\{s_1, s_2, ..., s_{N\times N}\right\}$ by:
\begin{equation}
c_i = s_i \oplus k_i,
\end{equation}
where $1 \leq i \leq N\times N$ and $\oplus$ represents bitwise exclusive OR operation. So the encrypted sequence $C = \left\{c_1, c_2, ..., c_{N\times N}\right\}$ is Obtained.
\\(5) Obtained the cipher-image by reshaping the encrypted sequence $C$ into an $N \times N$ image.


\section{Proposed method}

To attack the chaos-based image encryption algorithm \cite{guan2005chaos}, we need to find a complex mapping function to model the inverse transform from encrypted images and plain images. We employ deep convolutional neural networks (CNN) \cite{lecun2015deep,krizhevsky2012imagenet} to model such complex inverse functions \cite{RevHashNet18, PyLRRNet19}.           

In Fig.\ref{model}, the model is mainly divided into convolutional groups and deconvolutional groups. In convolutional groups, the input are the cipher images described as $\mathbf{X}$, we design several convolutional layers to analyze input image composition and obtain the low-dimensional features, the operation is defined as $\mathbf{Y} = O(\mathbf{X})$. In deconvolutional groups, we perform the opposite operation to the convolution stage and reconstruct plain images with high fidelity. The inverting operation is described as $\mathbf{\tilde{X}} = H(\mathbf{Y})$. The regenerate images are compared to their ground truth plain images given as target $\mathbf{T}$, we use Mean Squared Error (MSE)
as the loss function \cite{ishikawa1996structural,dong2014learning,lotter2016deep,levine2016end}. After training, the model can be used to attack the chaos-based image encryption algorithms \cite{guan2005chaos}.
 

\subsection{Network Architecture}
\begin{figure}[ht!] 
\centering
\includegraphics[width=0.35\textwidth]{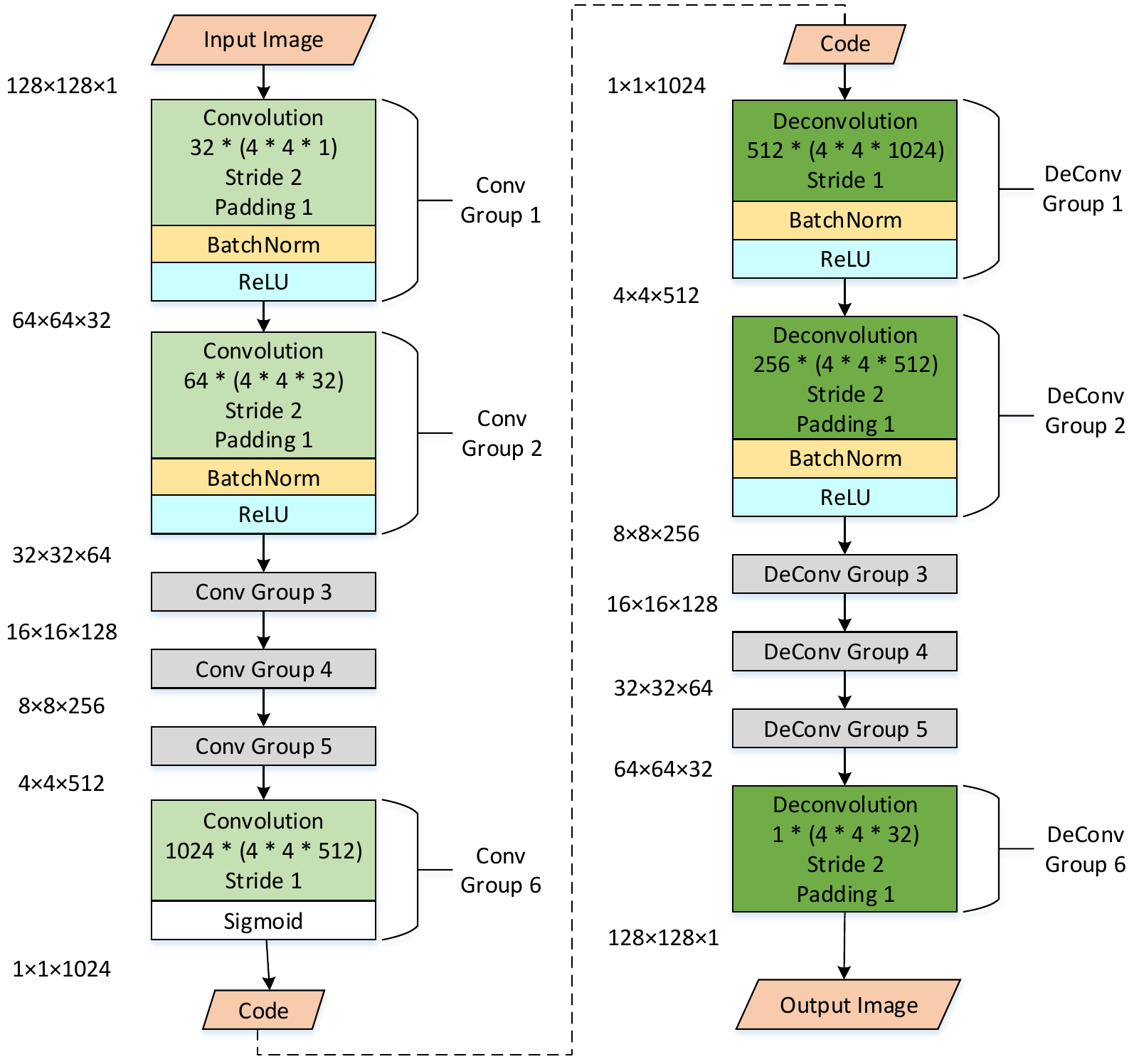} 
\caption{The network structure of the deep learning-based model, which is also mainly composed of the convolutional groups and deconvolution groups. The former mainly includes $6$ convolution layers and the latter mainly has $6$ deconvolution layers.}
\label{network}
\end{figure}

 As shown in Fig.\ref{network}, on the left is the convolutional groups and the deconvolutional network is on right, they are basically symmetrical. 

In convolutional groups $1$ to $5$, in order to ensure that the information at the edge of the image can be utilized, one-dimensional zero padding\cite{lecun2015deep} is first applied around the input image. Then the padded image is convolved with a convolutional layer ($4 \times 4$ kernel size, $2 \times 2$ stride). Next, we perform $Batch Normalization$ on the output \cite{ioffe2015batch}. Finally the output is put through the $relu$ \cite{glorot2011deep} function described as $relu(x) = \max(0, x)$, introducing non-linear factors to neurons. In convolutional group $6$, we only construct a convolutional layer ($4 \times 4$ kernel size, $1 \times 1$ stride) and  activation function $\sigma(x) = 1/(1+e^{-x})$\cite{lecun2015deep}.

In the convolutional groups above, except that the number of output feature maps of the convolutional layer in the first group is $32$, the output of other convolutional layers is twice that of the corresponding input feature map number.
For each $i(1\leq i\leq 6)$ of the groups, the convolutional operation $o_i$ is given by
\begin{equation}
      o_i( \mathbf{X})=
      \begin{cases}
            \mathbf{X}, & i=0 \\
            relu(\amalg(o_{i-1}(\mathbf{X}) \otimes\mathbf{W}_i + \mathbf{b}_i)), 
                & 1\leq i\leq 5 \\
            \sigma(o_{i-1}(\mathbf{X}) \otimes\mathbf{W}_i + \mathbf{b}_i),
                & i=6,
      \end{cases}
\end{equation}
where $\mathbf{W}_i$ and $\mathbf{b}_i$ denote the weights and biases of convolutional filters, respectively. $\otimes$ represents the convolution operator and $\amalg$ represents $Batch Normalization$.

For the convolutional groups, there are six deconvolutional groups. The first deconvolutional group includes a deconvolutional layer ($4 \times 4$ kernel size, $1 \times 1$ stride) followed by a $Batch Normalization$ layer and a $relu$ function. In deconvolutional groups $2$ to $6$, we design a deconvolutional layer ($4 \times 4$ kernel size, $2 \times 2$ stride) with one-dimensional zero padding, followed by a $Batch Normalization$ layer and $relu$ function except group 6.

In the above deconvolutional groups, the number of output feature maps for the deconvolution layer in the last group is one, while the number of output feature maps for other deconvolution layers is half of the corresponding input. For a single group $k(1\leq k\leq 6)$, given the deconvolution operator $\odot$, deconvolutional filter weights $\mathbf{\tilde{W}}_k$ and biases $\mathbf{\tilde{b}}_k$, the deconvolutional operation $h_k$ is described as   
\begin{equation}
      h_k(\mathbf{Y})= 
      \begin{cases}
             \mathbf{Y}, & k=0 \\
             relu(\amalg(h_{k-1}(\mathbf{Y}) \odot\mathbf{\tilde{W}}_k + \mathbf{\tilde{b}}_k)), 
                & 1\leq k\leq 5 \\
             h_{k-1}(\mathbf{Y}) \odot\mathbf{\tilde{W}}_k + \mathbf{\tilde{b}}_k,
                & k=6.
    \end{cases}
\end{equation}

In the network designed above, the reason for using the convolution layer instead of the downsampling layer when projecting input into a low-dimensional space is that the downsampling layer loses more original information. In addition, the $4 \times 4$ convolution kernel size is used to simply and efficiently project the input into the low-dimensional space and to ensure the symmetry of the projected and reconstructed plain image.


\subsection{Network Training}
In order to model the inverse transform $H(O(\mathbf{X}))$ from encrypted images and plain images, we will minimize the distance between the output of the network and the plain images $\mathbf{T}$ corresponds to the input cipher images $\mathbf{X}$ by using MSE. In addition, we introduce the regularization of weight decay to help for a better generalization in order to avoid overfitting\cite{moody1992effective,sarle1996stopped,loshchilov2017fixing}. Given the regularization weighting coefficients $\alpha = 0.01$, we have loss function 
\begin{equation}
    L(\mathbf{W}, \mathbf{b})
   = \left\|\mathbf{T} - H(O(\mathbf{X}))\right\|^2_2
   + \alpha \left\| \mathbf{W} \right\|^2_2,
\end{equation}
where 
\begin{equation}
  \left\{
    \begin{array}{lr}
           \mathbf{W} = \sum\limits_{i} \mathbf{W}_i 
                      + \sum\limits_{k} \mathbf{\tilde{W}}_K \\ 
           \mathbf{b} = \sum\limits_{i} \mathbf{b}_i 
                      + \sum\limits_{k} \mathbf{\tilde{b}}_K.         
    \end{array}
  \right.
\end{equation}

After defining the loss function, we use the Adam \cite{kingma2014adam} optimization method to train it. While iteratively inputting each data batch ($batch\_size = 64$), the filter is updated in the direction that minimizes the loss function. We set the initial learning rate to $0.001$, and the coefficients used for computing running averages of gradient and its square is $(0.9, 0.999)$.


\section{Experiments}


\subsection{Cracking of images encrypted with static-keys} \label{static_sub}
In the static-key case, the MNIST dataset is used for training and cracking evaluations. In detail, we use $60,000$ plain images from the training set of the MNIST dataset, and the rest $10,000$ plain images from the test set. The precision is set as $10^{-14}$ when encrypting plain images using the chaos-based image encryption algorithm \cite{guan2005chaos}. The secret keys of Arnold's cat map are chosen as: $p=4$, $q=7$, $n=5$; the parameters of Chen's chaotic system are selected as: $a=35$, $b=3$, $c=28$. And the initial condition of Chen's system is: $x_0=-10.058$,  $y_0=0.368$, $z_0=37.368$. For details of selecting the secret parameters, please refer to \cite{guan2005chaos}. Next we resize all images into $128 \times 128$ pixels and store in PNG format. Finally, we use the resulting images (i.e., $60,000$ pairs of cipher-plain images) from the training set to train the network. To test the decryption performance, we feed the encrypted images from the test set (i.e., $10,000$ encrypted images) to the well-trained network.

\begin{figure}[ht!] 
\centering
\includegraphics[width=0.45\textwidth]{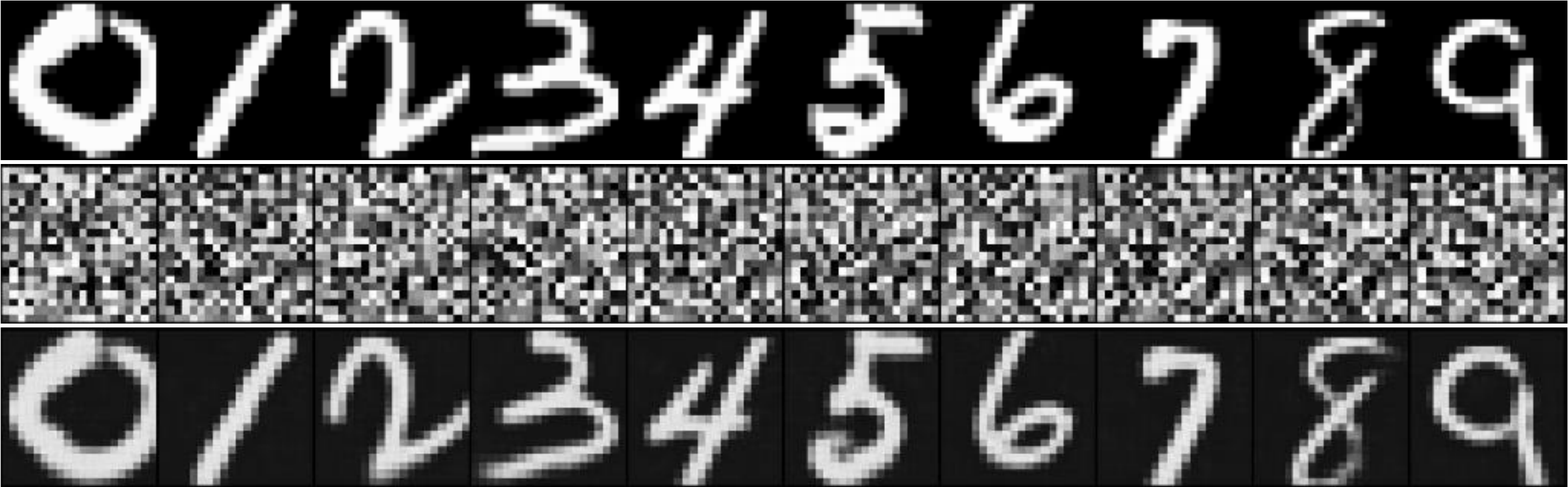} 
\caption{Cracking of images encrypted with static keys. From the top to bottom are the plain-images, cipher-images and regenerate images, and the regenerate images have a very high degree of reduction on the plain-images, we can clearly distinguish each number. }
\label{static_key}
\end{figure}

In Fig. \ref{static_key}, we show several decryption representatives. The top row shows ten plain digits which were randomly selected from the MNIST ten-digit categories. Corresponding to each digit sample in the first row, their cipher and regenerated images are shown in the second and the third row, respectively. In the second row, we find each digit has been randomly largely shuffled and changed with the chaos-based encryption \cite{guan2005chaos}. Comparing the regenerated images with the cipher images, we observe that the regenerated images mostly restored the content information of their plain images. One could clearly tell the digit number represented by each regenerated image.

\begin{figure}[ht!] 
\centering
\includegraphics[width=0.45\textwidth]{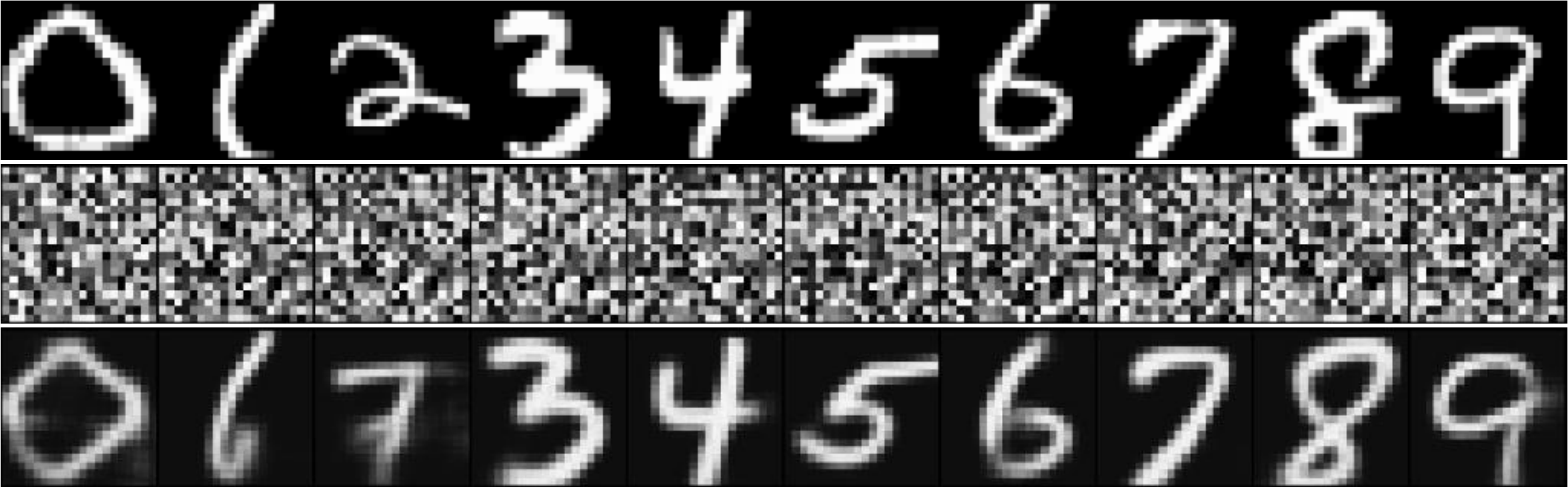} 
\caption{Cracking result of images encrypted with dynamic-keys. From the top to bottom are the plain-images, cipher-images and regenerate images, and regenerate-images still restores most of the information in the plain-images, and we could easily distinguish the numbers on each image, except for individual numbers such as $2$.}
\label{dynamic_key}
\end{figure}

\subsection{Cracking of images encrypted with dynamic-keys} \label{dynamic_sub}
In this experiment, we evaluate the image cracking performance in the dynamic-key setting. Specifically, we change the selection of secret parameters of Arnold's cat map by setting $p$ and $q$ from  a range $1$ to $9$. During encryption process, $p$ and $q$ will randomly select the parameters within this given range. All other secret keys settings remain the same as the static-key experimental settings in subsection \ref{static_sub}. In order to increase the coverage of our network for different encryption transformation methods due to distinct keys, we encrypted each sample of the MNIST training set four times using a set of dynamic-keys. With this, we have $240,000$ pairs of cipher-plain images to train the network. Similarly, the test set is encrypted with dynamic keys, and test results are shown in Fig. \ref{dynamic_key}.

In the experimental results Fig.\ref{dynamic_key}, plain-images, cipher-images and regenerated images are also arranged in order from top to bottom. The regenerated images still restore most essential information of the plain-images but there are also a few collapsed results, e.g., the restored image of the number $2$ in the Fig.\ref{dynamic_key}. Despite several failed cases, we could still easily distinguish the numbers on the images except these individual examples. 


\section{Evaluations}
\subsection{Quantitative Evaluations}

In order to measure the quality of our experimental results, we introduce the $LeNet-5$ network \cite{lecun1998gradient} to classify and evaluate the results, because the $LeNet-5$ network can classify MNIST data sets with high precision, its accuracy on the MNIST test set can reach $ 99.05\% $, and its structure is relatively simple, so it is very efficient to use it as an experimental evaluation. In Evaluations, we first tested the classification accuracy of $LeNet-5$ on the original MNIST test set, and then used it to evaluate the performance of our decrypted images. Table \ref{tab:acc} shows the classification accuracy rates of $LeNet-5$ network on MNIST test set and our experimental results. Among them, the accuracy rate on the original MNIST test dataset is $98.77\%$, which indicates the efficiency of $LeNet-5$ for the MNIST digit recognition. To quantitatively test the accuracy of our pre-trained decryption model, firstly we encrypt the MNIST test dataset using the chaos-based encryption \cite{guan2005chaos} in the static-key encryption scenario. Then the encrypted MNIST digits are decrypted using our pre-trained model to approximately decrypt plain images. Finally, the decrypted images are recognized by the $LeNet-5$ model. The accuracy is calculated by comparing the predicted label from decrypted images with ground truth label from their plain image counterparts. The recognition accuracy is $97.87\%$, which is only slightly lower than that on the plain images. This means that decrypted images can be recognized quite well. Similarly, we test the recognition accuracy of decrypted images from the dynamic-key encryption. As showed in Table \ref{tab:acc}, the accuracy rate is $92.04\%$. From the recognition results, we conclude that the proposed image decryption method can efficiently crack the chaos-based image encryption methods \cite{guan2005chaos}.       

\begin{table}[ht] 
\caption{Classification accuracy rates of LeNet-5} 
\centering 
\begin{tabular}{|c|c|c|c|} 
\hline 
 \multirow{2}*{Image Set} & \multirow{2}*{Encryption} & Encryption  & Classification \\ 
 & & case & accuracy rates\\
\hline 
origin MNIST  & no & none & 98.77\% \\ 

\hline 
decrypted MNIST  & yes & static-key & 97.87\% \\ 

\hline 
decrypted MNIST  & yes & dynamic-key & 92.04\% \\  

\hline  
\end{tabular} 
\label{tab:acc} 
\end{table} 

\subsection{Comparison}
The attack method proposed in \cite{cokal2009cryptanalysis} is a process of manually analyzing the encryption algorithm \cite{guan2005chaos}. Among them, the chosen-plaintext attack requires constructing a specific plaintext input into the encryption algorithm\cite{guan2005chaos}, but many times there is no such condition. In contrast, our approach does not require specific inputs. And for the known-plaintext attack in \cite{cokal2009cryptanalysis}, the secret parameter extraction of Arnold's cat map is a process of continuously narrowing the scope of encryption transformation, the attacker needs to constantly find out all the possible encryption transformation methods until the final private parameters are found, which is an extremely time-consuming and labor-intensive process. In comparison, our method has the following advantages. The decryption method in \cite{cokal2009cryptanalysis} needs to be calculated separately for each crack. However, our method does not require to crack the keys before decryption, which is key-independent. For instance, in the dynamic-key encryption case,it only requires to be trained once to automatically decrypt MNIST digits encrypted with different keys. The experimental results confirm that the proposed method has certain degree of generalization ability.   


\section{Conclusion}

In this letter, we first present a new attack method based on deep learning for a chaos-based image encryption algorithm \cite{guan2005chaos}. The proposed method first projects encrypted images to the low-dimensional feature space. Then decrypted images are perceptually reconstructed with the deconvolutional generator. Experimental results verify the decryption accuracy of the proposed method in both static-key and dynamic-key encryption cases. Compared with the previous method, our solution is key-independent and automatical. In the future work, we will explore the feasibility of decrypting chaos-based video encryption and other image encryption schemes.   

\bibliographystyle{IEEEtran}
\bibliography{ref}

\end{document}